\documentclass[journal]{IEEEtran}
\usepackage{amsmath}
\usepackage{multirow}
\usepackage[table,xcdraw]{xcolor}
\usepackage[normalem]{ulem}
\useunder{\uline}{\ul}{}
\usepackage{float}
\usepackage{cite}
\usepackage{url}
\usepackage{times} 
\usepackage[numbers,sort&compress]{natbib}

\usepackage{blindtext}
\usepackage{gensymb}
\usepackage{enumitem}
\usepackage{academicons}
\usepackage{hyperref}

\ifCLASSINFOpdf
   \usepackage[pdftex]{graphicx}
\else
\fi
\newcommand{\orcid}[1]{\href{https://orcid.org/#1}{\textcolor[HTML]{A6CE39}{\aiOrcid}}}

\begin{document}
%
\title{Model-based Evaluation of Driver Control Workloads in Haptic-based Driver Assistance Systems}
%
%
%

\author{Kenechukwu C. Mbanisi \href{https://orcid.org/0000-0001-5871-3730}{\includegraphics[scale=0.15]{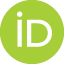}}, Hideyuki Kimpara \href{https://orcid.org/0000-0001-6441-6062}{\includegraphics[scale=0.15]{figures/orcid.png}}, \textit{Member, IEEE}, Zhi Li
\href{https://orcid.org/0000-0002-7627-947X}{\includegraphics[scale=0.15]{figures/orcid.png}}, \textit{Member, IEEE}, Danil Prokhorov, \textit{Senior Member, IEEE}, and Michael A. Gennert,
\href{https://orcid.org/0000-0002-3170-2190}{\includegraphics[scale=0.15]{figures/orcid.png}},
\textit{Senior Member, IEEE}


\thanks{Kenechukwu C. Mbanisi and Michael A. Gennert were with the Robotics Engineering Department (RBE), Worcester Polytechnic Institute (WPI), Worcester, MA 01609, USA, while Zhi Li is with WPI RBE (email: {kcmbanisi, michaelg, zli11}@wpi.edu)}

\thanks{Hideyuki Kimpara was with Toyota Motor North America (TMNA) R\&D, Ann Arbor, MI 48105 USA, and WPI RBE (e-mail: hideyukikimpara@gmail.com), and Danil Prokhorov is with TMNA R\&D (e-mail: danil.prokhorov@toyota.com)}

\thanks{This work was supported in part by Toyota Motor North America.}
}

%
%

\markboth{Journal of \LaTeX\ Class Files,~Vol.~XX, No.~X, October~20XX}%
{Mbanisi \MakeLowercase{\textit{et al.}}: Model-based Evaluation of Driver Control Workloads in Haptic-based Driver Assistance Systems}
%



\maketitle

\begin{abstract}
This study presents a novel approach for modeling and simulating human-vehicle interactions in order to examine the effects of automated driving systems (ADS) on driving performance and driver control workload. Existing driver-ADS interaction studies have relied on simulated or real-world human driver experiments that are limited in providing objective evaluation of the dynamic interactions and control workloads on the driver. Our approach leverages an integrated human model-based active driving system (HuMADS) to simulate the dynamic interaction between the driver model and the haptic-based ADS during a vehicle overtaking task. Two driver arm-steering models were developed for both tense and relaxed human driver conditions and validated against experimental data. We conducted a simulation study to evaluate the effects of three different haptic shared control conditions (based on the presence and type of control conflict) on overtaking task performance and driver workloads. We found that No Conflict shared control scenarios result in improved driving performance and reduced control workloads, while Conflict scenarios result in unsafe maneuvers and increased workloads. These findings, which are consistent with experimental studies, demonstrate the potential for our approach to improving future ADS design for safer driver assistance systems.
\end{abstract}


%
\IEEEpeerreviewmaketitle

\section{Introduction}
%
%
%
%
\IEEEPARstart{A}{utomated} driving systems (ADS) are a set of technologies with the potential to improve vehicle safety and reduce fatal crashes due to human error~\cite{campbell2018human}. 
A taxonomy of driving automation levels defined by SAE International (SAE J3016)~\cite{J3016_202104}, indicates that with higher autonomy levels, the human transfers increasing levels of driving responsibility and control authority to the automated driving system.
Despite remarkable advances in recent times, ADS have yet to reach the level of robustness and reliability necessary for widespread deployment. Therefore, systems that provide driver assistance while keeping the human driver in the loop (Level 1 \& 2) remain crucial for driver and road safety.

\begin{figure}[h]     
    \centering
    \includegraphics[width=1\columnwidth]{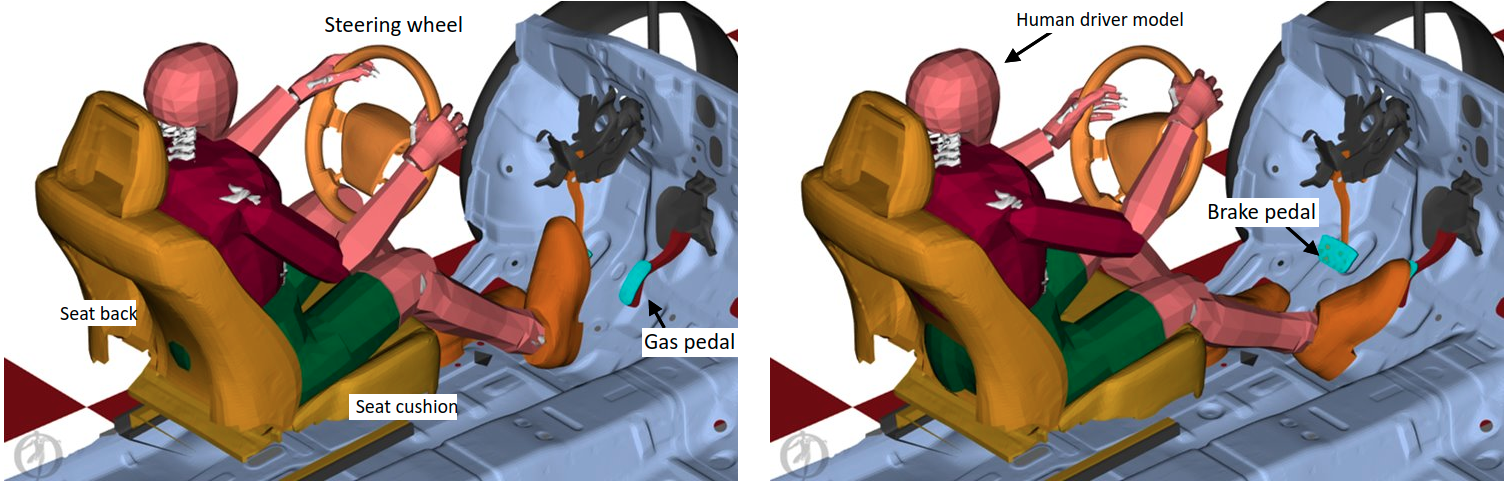}
    \caption{Mechanical models of the human model-based active driving system (HuMADS) on OpenSim simulation platform, (left) steering angle zero w/ pressing brake pedal, and (right) steering wheel angle 26 degrees w/pressing gas pedal.}
    \label{fig:drivermodel}
\end{figure}

Haptic-based design of levels 1 and 2 driving autonomy enables the driver and the ADS to cooperatively control the vehicle by blending their control inputs over a haptic control interface (e.g., a motorized steering wheel). 
The continuous haptic feedback allows the driver to be constantly informed of the ADS's actions and ensures they retain final control authority over the vehicle~\cite{Abbink2018}. Haptic shared control (HSC) has been applied to lane-keeping assistance~\cite{abbink2012haptic}, car-following support~\cite{mulder2010design}, collision avoidance~\cite{balachandran2015predictive} and has been shown to improve driving performance, situation awareness, and response times in driving tasks.

When conflict arises between the driver and the ADS, however, these systems result in increased interaction forces. In control conflicts, mismatches in situation awareness, decision making, and task execution can arise between agents~\cite{Itoh2016} or due to a mismatch in the tuning of haptic forces in response to the estimates of the dynamic response of the driver~\cite{abbink2012importance}. This often results in higher force loading on the driver, which contributes to increased physical workload and lowered user acceptance~\cite{Itoh2016}. The goal of this work is to evaluate the impact of HSC designs on driver control workload and physical stress in order to inform the design of safer haptic-based driver assistance systems. Generally, studies of co-driving systems use one of two existing approaches. Most studies rely on subjective evaluations of human subjects in simulated driving experiments to assess workload measures~\cite{hosseini2016predictive, de2014effects}. Alternatively, some studies use suitably instrumented human subjects and/or vehicle control devices in driving studies to identify force interactions and the resulting internal dynamics of the human driver~\cite{wang2020adaptive, nacpil2021application}.

We propose a third approach. That is, we use surrogate human driver models, as they enable accurate simulation of human behavior and dynamic characteristics, as well as providing access to evaluate hard-to-measure internal workload variables. Surrogate models have been used to evaluate the ergonomics of human-robot interactions~\cite{maurice2016experimental} and to determine the effectiveness of steer-by-wire driving systems~\cite{mehrabi2015steering}.

In this paper, we present a novel simulation framework for assessing the effects of HSC designs on the estimated driver control workload, force-torque interactions and driving performance. Our current work utilizes the human model-based active driving system (HuMADS) reported in our previous work~\cite{kimpara2019human}. This model, which is based on the OpenSim biomechanics platform~\cite{seth2018opensim} and SimBody numerical physics solver~\cite{sherman2011simbody}, enables simulation of human driving behavior based on driver behavior modeling, robotics theory, and vehicle dynamics. Fig.~\ref{fig:drivermodel} shows snapshots of HuMADS in action. In~\cite{kimpara2019human}, HuMADS demonstrated human driver maneuvers and driving performances, and validated them against previously published highway driving data~\cite{saigo2013investigation} in terms of vehicle velocity and  trajectory. In this paper, our main contributions are: 
\begin{enumerate}[leftmargin=*]
    \item Simulation framework that integrates human driver model and ADS to quantify driver control workload by accurately simulating human behavior and dynamic characteristics. 
    \item Definition of two simulated driver model modes (tense and relaxed) that correspond to the equivalent human arm-steering impedance characteristics in literature~\cite{pick2007dynamic}.
    \item Generation of realistic vehicle overtaking maneuvers in a HSC scenario between HuMADS and the ADS. 
    \item A simulation study to evaluate driving performance and driver control workloads across various HSC conditions.
\end{enumerate}

The structure of this article is as follows: Section II presents the system structure of HuMADS. Section III and Section IV show the validation of human arm impedance and performance of the overtaking driving task using the HuMADS and various HSC conditions. Section V and VI present the results and discussion of the simulation study. Limitations and conclusions are presented in Sections VII and VIII, respectively.

\section{Methods of Human Model-based Active Driving }  

\begin{figure*}[th]
        \centering
        \includegraphics[width=1.75\columnwidth]{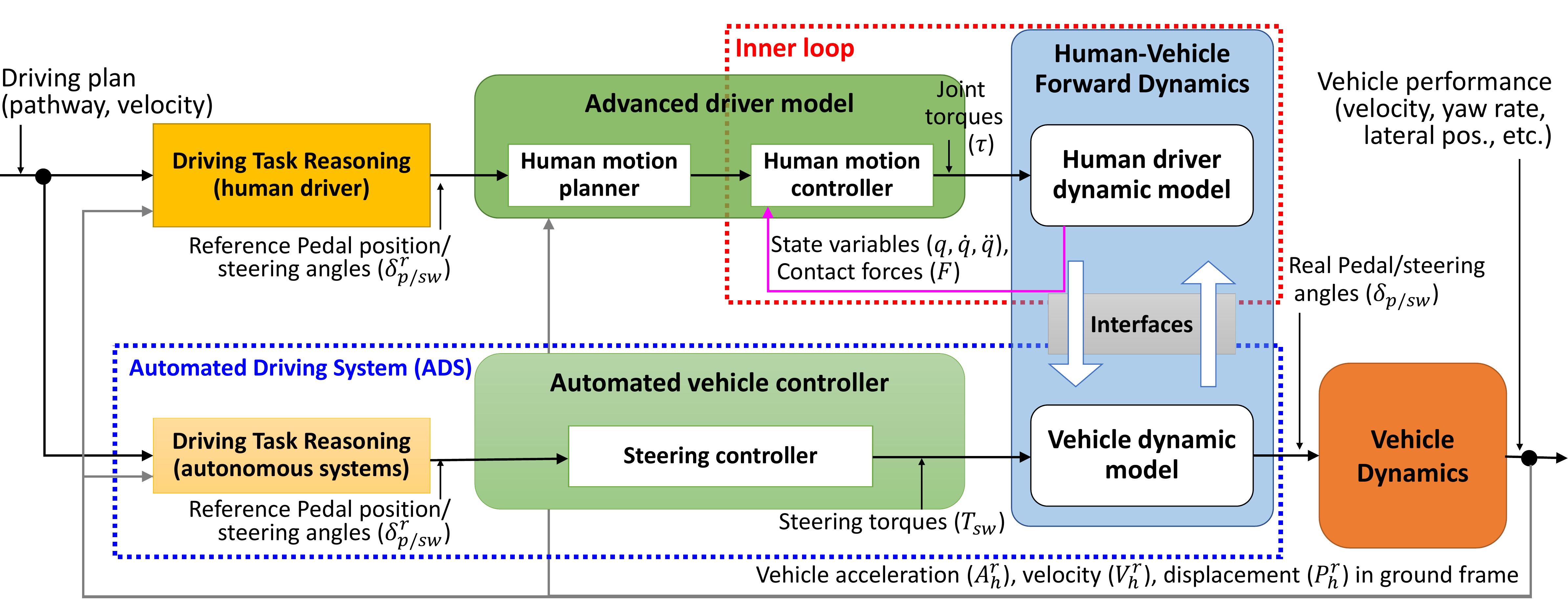}
        \caption{A simulation framework of the Human model-based active driving system (HuMADS) for human driver-vehicle interactions with the automated driving systems (ADS) in parallel.}
        \label{fig:combFramework}        
\end{figure*}

The simulation framework of the HuMADS for advanced driver-vehicle interaction (DVI) study comprises a closed loop feedback system with four major functions as shown in Fig.~\ref{fig:combFramework}. This framework has been designed to perform a car-following task using pedal and steering maneuvers of human drivers. 
Interactions between human and vehicle models are made up of outer and inner feedback control loops. In the outer loop, the \textbf{driving task reasoning} and \textbf{vehicle dynamics} control the longitudinal and lateral motions of the vehicle (i.e. accelerating, breaking, following lanes, etc.). While in the inner loop, the \textbf{human motion controller} controls the dynamic human driver behavior.
The driving task reasoning and vehicle dynamics are designed based on a linear feedback system proposed by Saigo~\cite{saigo2013investigation, raksincharoensak2014driver}.  The advanced driver model consists of a \textbf{human motion planner}, \textbf{human motion controller}, and a part of \textbf{human-vehicle forward dynamics} analysis. In addition to the advanced driver model, this study adds an \textit{\textbf{automated vehicle controller}} as an intelligent driving system to actuate vehicle control instruments directly. Physical interaction and force equilibrium analysis are performed on the dynamic models in the human-vehicle forward dynamics analysis based on the physics solver, SimBody~\cite{sherman2011simbody}. Because human body models and vehicle interior structures are expressed in the framework, it can be used to investigate human safety and ergonomics based on the design of vehicle interior and functional performance. The following subsections will briefly describe each component of the HuMADS framework. The full system description of the simulation framework was outlined in our previous publication~\cite{kimpara2019human}.

\subsection{Driving Task Reasoning}

The driving task reasoning layer decomposes control of longitudinal, i.e. forward motion, from lateral, i.e. steering motion. \textit{Longitudinal motion} is governed by the accelerator pedal and its reference position $\delta_{p}^{r}$. The longitudinal motion controller is based on a car-following model following Saigo's model~\cite{saigo2013investigation, raksincharoensak2014driver}. The inputs to the model are the velocity of the preceding vehicle and the road pathway. The reference pedal angles $\delta_{p}^{r} $ can be described by the following control equation:

\begin{equation}
    \delta_{p}^{r} = H_{D}(D_{hw} - D_{hw}^{d}) + H_{V}(V_{p} - V_{h}^{r})
\end{equation}

\noindent
where $H_{D}$ and $H_{V}$ are the gain constants for headway distance and vehicle velocity, respectively,  $D_{hw}$ and $D_{hw}^{d}$ are the actual and desired headway distances between the preceding vehicle and host vehicle. $V_{p}$ is the velocity of the preceding vehicle while $V_{h}^{r}$ is the reference velocity of the host vehicle.


\textit{Lateral motion} is governed by the steering wheel and its reference position $\delta_{sw}^{r}$. The lateral motion controller is a forward-gaze steering model for straight-road highway scenarios based on the theory of vehicle dynamics described by Abe~\cite{abe2015vehicle}. The reference steering angle $\delta_{sw}^{r}$ can be described by the following:

\begin{equation}
    \delta_{sw}^{r}(t) = hy_{srm}
\end{equation}

\noindent
where $h$ is a gain for the driver corrective steering angle and $y_{srm}$ is the lateral deflection between the desired road pathway point and the actual projected vehicle position based on the gaze model. The reference steering angle $\delta_{sw}^{r}$ is achieved by the inner-loop motion planner and controller, described in the advanced human driver model and human-vehicle forward dynamics functions. 
Note that the HuMADS framework is agnostic to specific driver task reasoning models (e.g. car-following, or forward-gaze model used in this study), hence other methods (e.g. see review paper~\cite{marcano2020review}) can be applied to simulate different driving behaviors.

\subsection{Human Motion Planner}

\begin{figure}[]
    \centering
    \includegraphics[width=0.8\columnwidth]{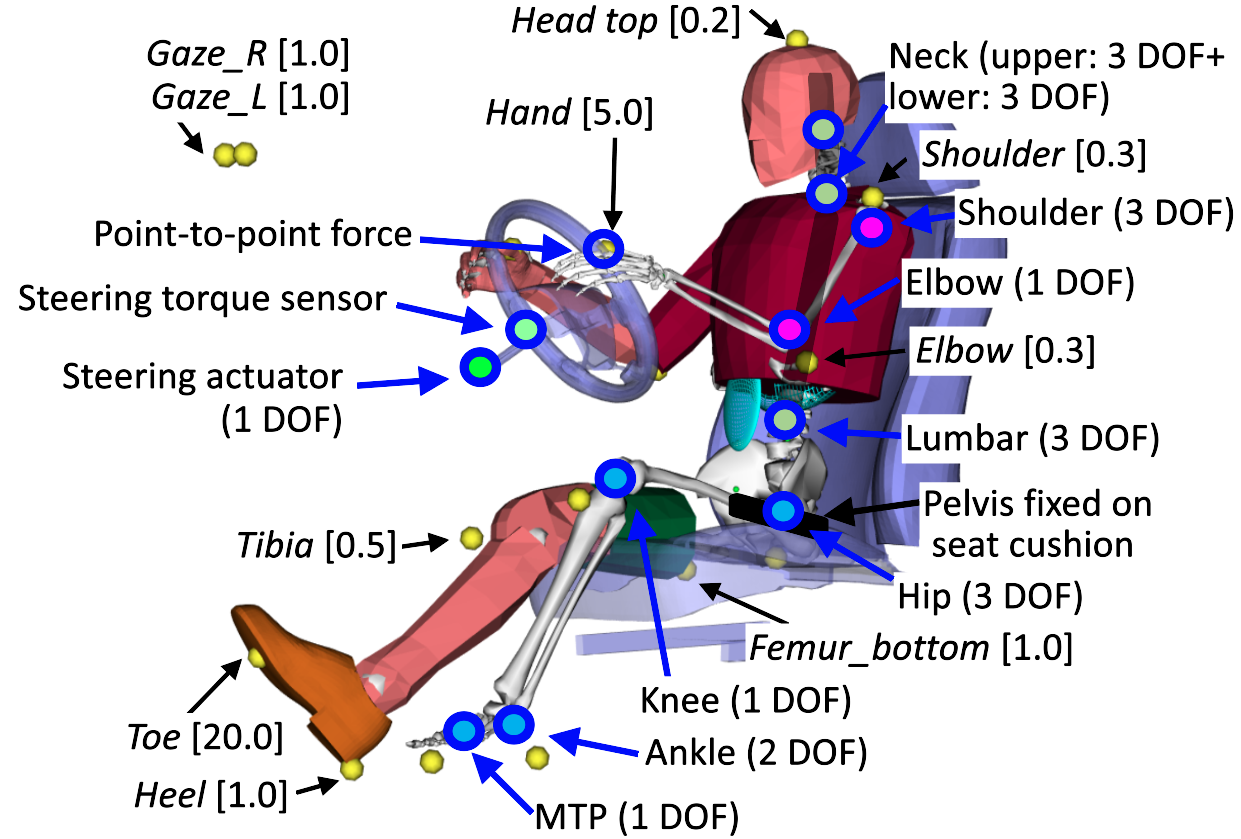}
    \caption{Virtual markers for inverse kinematics (IK) analysis and joints of human driver model. Names of virtual markers are written in italics with weights in [square brackets].  Degrees of freedoms (DOF) of joints are indicated in (parentheses).}
    \label{fig:virtualmarkers}
\end{figure}

The human motion planner, as part of the advanced human driver model, computes the desired joint angles $q^d$, angular velocities $\dot{q}^d$, and angular accelerations $\Ddot{q}^d$ given the reference pedal angles $\delta_{p}^{r}$ and steering angle $\delta_{sw}^{r}$ that control the longitudinal and lateral motions of a vehicle. To obtain those joint space variables, we employed OpenSim’s inverse kinematics (IK)~\cite{seth2018opensim} to obtain desired joint angles from planned trajectories. Virtual markers are used in OpenSim to define a specific position in 3D cartesian space. OpenSim’s IK analysis computes the optimized postures using the weighted least squares method to fit a set of desired marker positions based on their respective weighting. The weights were determined by iterative tuning using the heuristic of assigning higher weight values to the model's extremities (i.e. hands, feet) which require higher positional accuracy. The selected weights (seen in Fig.~\ref{fig:virtualmarkers}) are consistent with our previous work~\cite{kimpara2019human}.

Seventeen virtual markers were placed on the human driver model. Markers were placed on the tip-end positions of the upper and lower extremities (i.e. Hand and Toe in Fig.~\ref{fig:virtualmarkers}) to enable steering and pedal pressing. Whereas, the other virtual markers such as Elbow, Shoulder, Heel, Tibia, Femur-bottom, Head top, and Gaze-R/L were placed to keep the human body model in driving positions.
Since the pelvis, which is a base for the hierarchical human body model, may not move significantly during lane change motions, all 6 DOF of pelvis body against seat cushion were locked in this study. 

Upper limbs have 3 DOF shoulder joints and 1 DOF elbow joint. During overtaking, the hands of human drivers are assumed to remain on the steering wheel, so the steering wheel and hands are connected by point-to-point force models. This would be equivalent to a 3 DOF rotational joint between hands and steering wheel.


\subsection{Human Motion Controller}

The human motion controller, as part of the advanced human driver model, computes the desired joint torques $\tau$ given the desired joint angles $q^d$, angular velocities $\dot{q}^d$, and angular accelerations $\Ddot{q}^d$ of the vehicle maneuver motion as well as external forces.  We compute the desired joint torque for the whole-body human driver model to maintain its body posture while tracking the desired trajectory for pedal pressing and steering through inverse dynamics~\cite{kimpara2019human}. 
Inverse dynamics~\cite{spong2020robot} adopts the feedback linearization approach to express the joint torques $\tau$ as the linear function of model states and their derivatives along with external forces, $f(q, \dot{q}, \Ddot{q}, F_{ext})$.
Therefore, the modified dynamics equation can be stated with an input $a_q$ as follows:

\begin{equation}
    \tau = M(q)a_q + C(q, \dot{q}) + N(q) + J^TF_{ext}
\end{equation}

\noindent
where $q$ is the state variables in joint space, $M$ is the mass matrix,  $C$ is the velocity vector which is the product of the Coriolis and Centrifugal force matrix, and $\dot{q}$ is the velocity of the state variables, $N$ is the vector of torques due to gravity, $J^T$ represents the system Jacobian matrix, $F_{ext}$ captures the external forces (contact forces between the human model and the vehicle such as thigh and seat cushion forces), and $\tau$ is the joint torque vector. 

Given the desired joint positions, velocities, accelerations, and assuming that the human driver may have desired reaction forces on hands and feet during driving given that drivers have higher vibration sensitivity on hands and feet~\cite{campbell2018human}, our motion controller guarantees the following control law:

\begin{equation}
    \Ddot{\hat{q}} + K_D\dot{\hat{q}} + K_P\hat{q} + K_FJ_b^T\hat{F}_{rc} = 0
\end{equation}



\noindent
where $q^d$ is the vector of desired state variables, $\hat{q}$ represents the error of state variables ($\hat{q} = q - q^d$, $\dot{\hat{q}} = \dot{q} - \dot{q}^d$, $\Ddot{\hat{q}} = \Ddot{q} - \Ddot{q}^d$), $K_P$ and $K_D$ are diagonal matrices of position and velocity gains, respectively, $\hat{F}_{rc}$ is the error of the reaction forces $(\hat{F}_{rc} = F_{rc} - F_{rc}^{d})$ on joint $b$ which sustains contact forces and $K_F$ is a diagonal matrix of force gains.  

To control a linear second-order system, we set $\Ddot{q}$ as the controlled acceleration $a_q$ in (5) such that:

\begin{equation}
    a_q = \Ddot{q}^d - K_D(\dot{q} - \dot{q}^d) - K_P(q - q^d) -  K_FJ_b^T(F_{rc} - F_{rc}^{d})
\end{equation}


Furthermore, we assume that the human driver will push (or pull) the steering wheel to align the current steering angle with the reference steering angle $\delta_{sw}^{r}$ obtained from the driving task reasoning layer, therefore, the desired reaction force on the driver's hands is expressed as follows:

\begin{equation}
    F_{rc}^{d} = -K_{GS}(\delta_{sw} - \delta_{sw}^{r})
\end{equation}

\noindent
where $K_{GS}$ is a steering angle gain.

\subsection{Human-Vehicle Forward Dynamics }

The human-vehicle forward dynamics layer uses the whole-body human driver and vehicle models to simulate the human driver's forward kinematics and dynamics considering inertia and interactive forces to calculate the actual control angles of the pedal $\delta_{p}$ and steering wheel $\delta_{sw}$. The HuMADS is designed to simulate typical driving scenarios, thus, the human driver model is represented as a mid-sized male with height of 1.8 m and weight of 77.4 kg driving a regular passenger car. Vehicle geometry and inertial properties come from a FE model of a 2012 Toyota Camry~\cite{reichert2016validation, Sedan2016DevelopmentV}. The joint stiffness characteristics of the steering wheel $T_{sw}$ is adapted from~\cite{mehrabi2015steering} as follows:  

\begin{equation}
    T_{sw}^{d} = K_{sw}\delta_{sw} - C_{sw}\dot{\delta}_{sw}^{r}
\end{equation}

\noindent
where $K_{sw}$ and $C_{sw}$ are the stiffness and damping coefficients. The mass of steering wheel is obtained from mass parameter of FE model. To simulate hand grasping on the steering wheel, point-to-point spring force models with stiffness of 5 kN/m and damping coefficient of 100 Ns/m are used to constrain hand positions on the wheel. 

Pedal force characteristics against the displacement for gas and brake pedals agree with literature data~\cite{ho2015influence}. The contact interfaces between the human body parts and vehicular interior instruments and components, including the gas and brake pedals are also considered. In particular, the contact geometries of the human body come from skin mesh models of the Hybrid III 50th percentile male dummy~\cite{lstc}.  The contact forces are computed using the ElasticFoundationForce model~\cite{seth2018opensim}, while the contact parameters are determined according to the data of the pendulum foot impact in~\cite{manning1998dynamic}. 

\subsection{Vehicle Dynamics}

The vehicle dynamics converts the driver's commanded pedal angles $\delta_{p}$ and steering wheel $\delta_{sw}$ to vehicle dynamics.  Vehicular lateral motions are expressed based on the equivalent bicycle model as in~\cite{kimpara2019human}. The resulting vehicle motions are fed back to the driving task reasoning layer.

\begin{figure}[t]
    \centering
    \includegraphics[width=0.98\columnwidth]{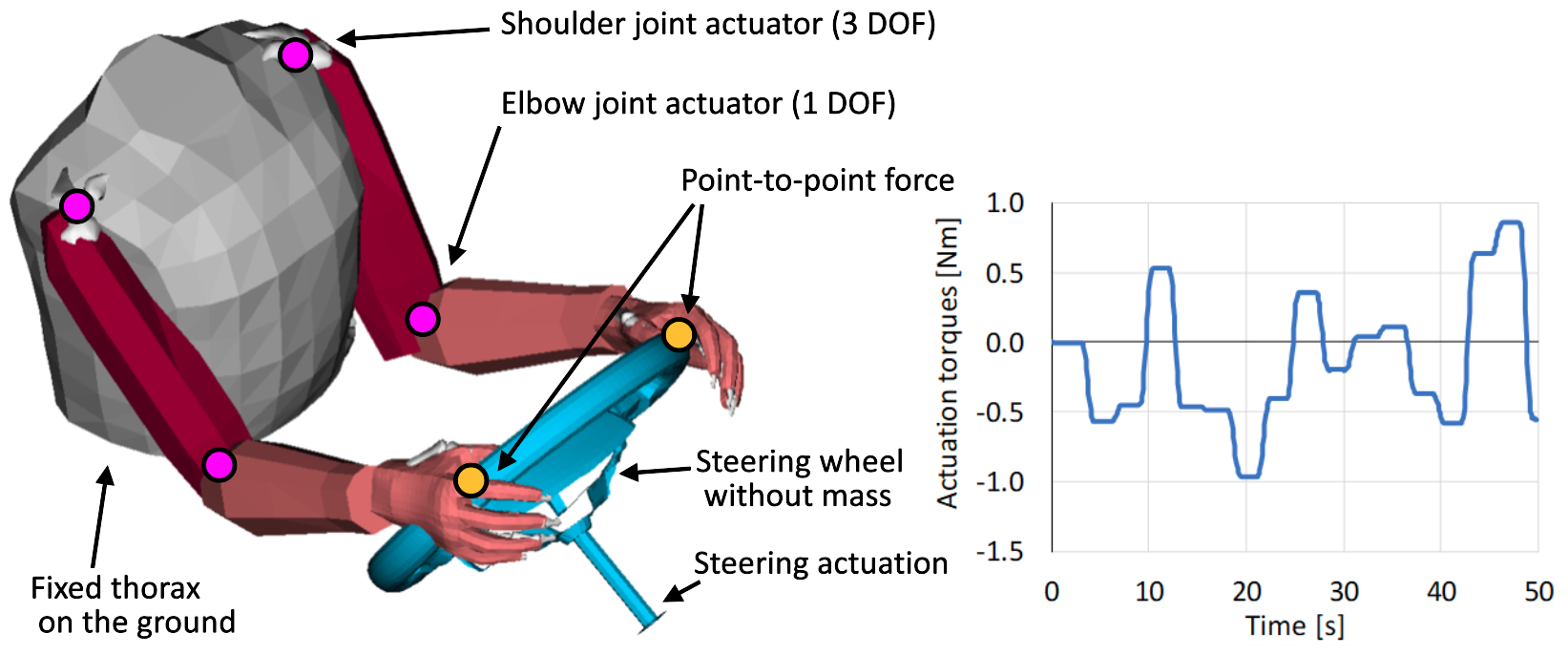}
    \caption{Simulation setup for equivalent impedance study and a pulse of applied steering actuation torque obtained from Mehrabi~\cite{mehrabi2014dynamics}}
    \label{fig:armsimsetup}
\end{figure}

\begin{table}[h]
    \centering
    \caption{Selected controller gains of driver model}
    \small
    \begin{tabular}{c||c||c}
    \hline
    & \bfseries Relaxed & \bfseries Tense \\
    \hline\hline
     $K_P$ & 30 & 225 \\ \hline
     $K_D$ & 10.8 & 30 \\ \hline
     $K_F$ & 0 & 1 \\ \hline
     $K_{GS}$ & 0 & 800 \\ 
    \hline
    \end{tabular}
    \vspace{1ex}
    \label{tab:controller_gains}
\end{table}

\section{Validation of Upper Limb Equivalent Impedance  }  

The equivalent impedance of the human driver's arms on the steering wheel are a crucial characteristic in describing the force equilibrium between human drivers and ADS actuations. Pick and Cole obtained equivalent impedance of human drivers holding a steering wheel from 8 human participants~\cite{pick2007dynamic}. They fit the measured data in relaxed and co-contracted arm conditions to the parameters of a mass-spring-damper model with a single degree of freedom as in the following equation:

\begin{equation}
    T_{mod}(s) = \frac{\theta(s)}{T_{hm}(s)} = \frac{1}{J_{hm}s^2 + B_{hm}s + K_{hm}}
\end{equation}

\noindent
where $T_{hm}$ is an equivalent steering torque due to human arm actuations, $K_{hm}$ and $B_{hm}$ are the torsional stiffness and damping of the driver’s arm as seen at the steering wheel, and $J_{hm}$ is the inertia of driver’s arm only.  

In our study, we estimate the equivalent arm-steering impedance characteristics of the driver-vehicle model using the mass-spring-damper model in (8). Specifically, we performed a validation study, following Mehrabi’s simulation approach~\cite{mehrabi2014dynamics}, where the driver model holds the steering wheel and we compute dynamic response of steering wheel against random actuated torque. Since our driver model is actuated via rotational joint actuators (rather than muscle elements), this study redefines the co-contracted case as tense case.

To achieve this, the whole-body model of HuMADS is simplified to a two-arm model with fixed thorax setup in order to represent same test configurations of~\cite{pick2007dynamic} as shown in Fig.~\ref{fig:armsimsetup}. Considering that the equivalent arm-steering model in (8) does not contain impedance parameters for the steering wheel, we model the steering wheel without mass and stiffness. The applied steering actuation torques are given from Mehrabi’s validation study~\cite{mehrabi2014dynamics}. 
Mehrabi adjusted the parameters associated with the passive properties of their driver model such as stretch reflex and joint stiffness~\cite{mehrabi2014dynamics}.  On the other hand, Beeman et al. reported that dynamic responses of human drivers are significantly different between relaxed human volunteer subjects and no-controlled human subject such as the post mortem human subjects (PMHS) from their sled test study~\cite{beeman2012occupant}. Because of their findings, we assume that human drivers may unconsciously utilize low-level controller for balancing even when asked to be fully relaxed. Since current HuMADS removed all muscle elements from original OpenSim human body models for simplicity, we decide to explore feedback gains of joint torques in equations (5) and (6) instead of applying passive mechanical parameters on the human joints. The desired joint angles $q^d$ are determined as the initial joint angles of holding steering wheel, and desired joint velocity $\dot{q}^d$ and acceleration $\Ddot{q}^d$ are set to zero. TABLE~\ref{tab:controller_gains} shows selected controller gains for relaxed and tense human driver arms. We assume that relaxed drivers are less likely to adjust their steering angle to the desired position. Thus, in relaxed driver mode, we set the force feedback gains in equations (5) and (6) to zero. 

\begin{figure}[t]
    \centering
    \includegraphics[width=0.75\columnwidth]{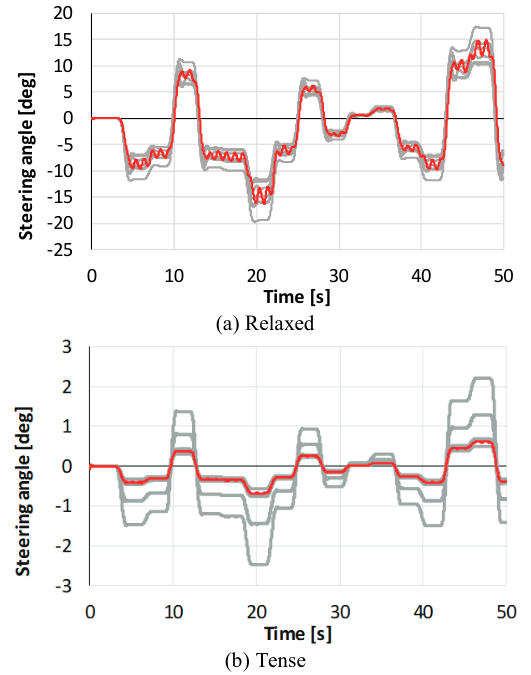}
    \caption{Comparison of steering wheel angle responses between identified human subject data~\cite{pick2007dynamic} (gray colored lines) and simulation results (red colored lines) with relaxed (a) and tense (b) modes.}
    \label{fig:validationresults}
\end{figure}

To validate our model-based approach, we compared the steering wheel angle response to a random steering torque between our model and experimental data. As seen in Fig.~\ref{fig:validationresults}, the steering wheel angle responses agree with the curves from human subject equivalent models in~\cite{pick2007dynamic} for both tense and relaxed modes. In the following section, we conduct a simulation study using the obtained controller gains to represent the relaxed and tense human driver modes.

\section{Overtaking Driving Task }  

\begin{figure}[]
    \centering
    \includegraphics[width=0.9\columnwidth]{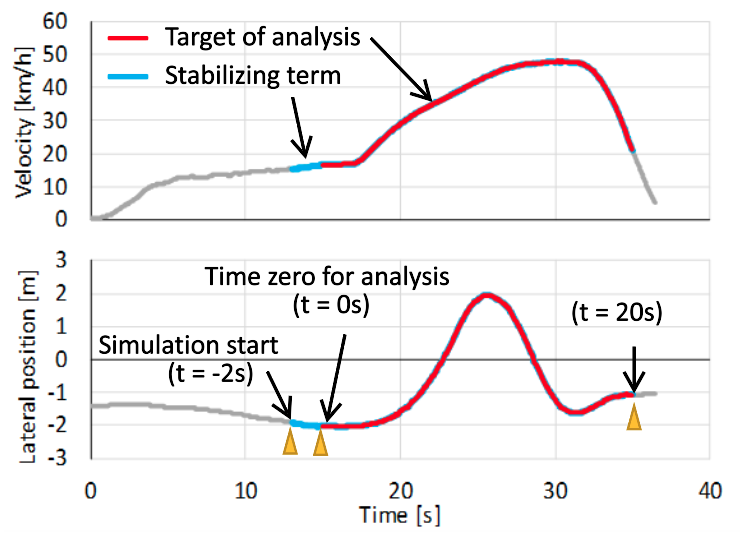}
    \caption{Vehicle input data of velocity and lateral position obtained from~\cite{naranjo2008lane}}
    \label{fig:vehiclesimulation}
\end{figure}

\begin{figure}[]
    \centering
    \includegraphics[width=0.9\columnwidth]{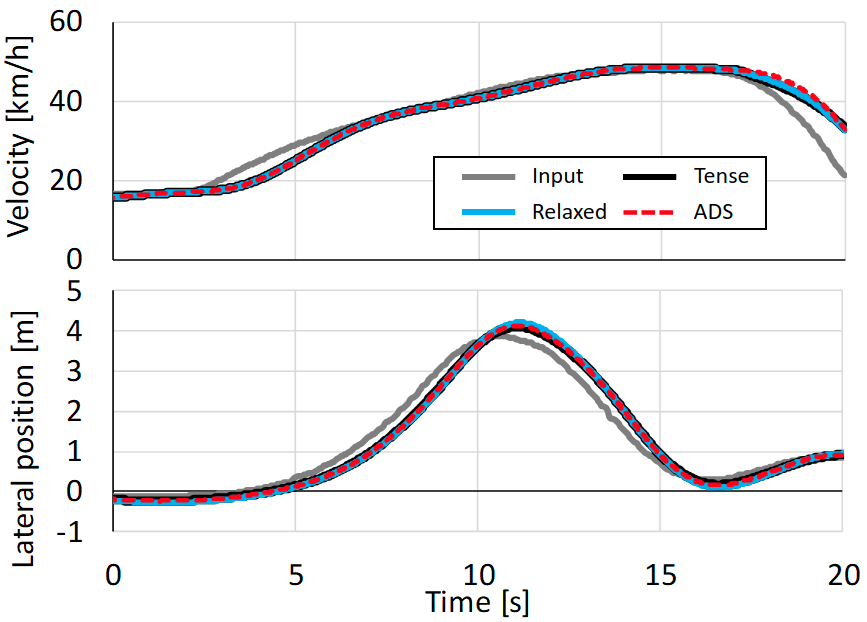}
    \caption{Comparisons of vehicle velocity and lateral position generated by HuMADS in tense and relaxed modes and ADS against input data.}
    \label{fig:vehiclesimresults}
\end{figure}

A vehicle overtaking maneuver involves the execution of two lane-change maneuvers. Toledo and Zohar reported that time durations for lane-changes around a highway traffic in California are 4.6$\pm$2.3 s and 4.4$\pm$2.4 s for directing towards the left and right, respectively~\cite{toledo2007modeling}.  In order to simulate the vehicle overtaking task on our HuMADS framework, we use the overtaking ego vehicle's motion data (i.e. vehicle speed and lateral displacement trajectories) from an experimental study~\cite{naranjo2008lane} as the desired inputs to our model.


\subsection{Framework for Shared Control with the Automated Driving System (ADS) }

In order to investigate a driver-ADS shared driving scenario, we integrate the ADS into the HuMADS framework as in Fig.~\ref{fig:combFramework}. The ADS must perform a perception-planning-action cycle alongside the human driver in order to contribute to the shared driving task. Consequently, this study defines a separate driving task reasoning and steering torque control for the ADS in parallel with that for the human driver model. As mentioned above, conflicts occur in haptic shared control when there is a mismatch in shared situation awareness, decision making or task execution between the driver and the ADS~\cite{Itoh2016}. 

In this paper, we consider two distinct cases of control conflict: (i) mismatch in the planned maneuver to take (Type I conflict) and (ii) mismatch in the resulting desired trajectory based on a planned maneuver (Type II conflict). In the context of an overtaking scenario, Type I conflict may occur if the driver wishes to perform an overtaking maneuver in the highway, however, the ADS chooses to remain in the current lane because of an oncoming vehicle in the blindspot of the driver. The Type II conflict instance may occur when both the driver and the ADS agree on the overtaking maneuver, however, with different resulting trajectories. This may occur as a result of a discrepancy between driver's behavior and the driver task reasoning model utilized by the ADS~\cite{boink2014understanding}.

The steering actuation torque $T_{ADS}$ of the ADS is calculated as follows: 

\begin{equation}
    T_{ADS} = -K_{GADS}(\delta_{sw} - \delta_{sw}^{r})
\end{equation}

\noindent
where $K_{GADS}$ is a steering angle gain for ADS actuator. In conflict scenarios, the driver may attempt to seize control over the vehicle by applying more torque to the steering wheel. We limit the maximum ADS actuator torque to 5 Nm in order to ensure that the driver can override the control torques of the ADS in conflict scenarios. A virtual steering torque sensor is inserted between the ADS actuator and steering wheel (see Fig.~\ref{fig:virtualmarkers}) to measure the resulting torque between the driver model and the ADS actuator.

\subsection{Simulation setup}


As mentioned above, we use the overtaking ego vehicle's motion data (i.e. vehicle velocity and lateral displacement trajectories) obtained from ~\cite{naranjo2008lane} as the input to our HuMADS model (see Fig.~\ref{fig:vehiclesimulation}). The simulation starts at 13 s of reference data, however, the first 2 s of time duration is used for stabilization of dynamics model because of discrepancies between initial and desired state variables.  The target of our analysis are the following 20 s of simulation. The entire simulation time (22 s) took a processing time of approximately 2.5 hours on a standard Windows PC with a single core, Intel i7 processor.

TABLE~\ref{tab:model_parameters} shows the parameters used for the driving task reasoning and steering wheel models. These parameter values as well as other model parameters for vehicle dynamics (such as vehicle mass, tire cornering forces, vehicle moment of inertia, wheelbase length, etc.) were set as in~\cite{kimpara2019human}.

\begin{table}[h]
    \centering
    \caption{Parameters for the driving task reasoning and steering wheel}
    \begin{tabular}{c c c}
    \hline\hline
    \bfseries Parameter for driving task reasoning & \bfseries Symbol & \bfseries Value \\
    \hline
     Desired headway distance behind preceding vehicle [s] & $T_{hw}^d$ & 0.5 \\ 
     Gain for distance difference [\%/m]  & $H_{D}$ & 20.0 \\ 
     Gain for velocity difference [\%/(m/s)]  & $H_{V}$ & 20.0 \\ 
     Time delay for human driver in pedal task [s]  & $\tau_{p}$ & 0.4 \\
     Steering compensation gain [rad/m]  & $h$ & 1.0 \\
     Forward-gaze time duration for steering task [s]  & $T_{p}$ & 0.5 \\
     Time delay in steering task [s]  & $\tau_{sw}$ & 0.1 \\
     
    \hline\hline
    \\
    
    \hline\hline
    \bfseries Parameter for steering wheel & \bfseries Symbol & \bfseries Value \\
    \hline
     Stiffness coefficient of steering wheel [N/rad] & $K_{sw}^d$ & 7.5 \\ 
     Damping coefficient of steering wheel [N s/rad]  & $C_{sw}$ & 0.9 \\ 
     Gain for ADS steering actuator [N m/rad]  & $K_{GADS}$ & 72.7 \\ 
     \hline\hline
     
    \end{tabular}
    \label{tab:model_parameters}
\end{table}

\subsection{Verification of Driving Performance }

We implemented the forward simulation on the HuMADS model in three modes: (1) Tense driver model only, (2) Relaxed driver model only, (3) ADS only. The model parameters for the tense and relaxed driver modes were based on our validation results in Table~\ref{tab:controller_gains}. Fig.~\ref{fig:vehiclesimresults} shows the simulation results generated by the three modes in comparison with the experimental input data. The maximum lateral error between the simulated and experimental data are 0.25, 0.35 and 0.27 m for the tense driver, relaxed driver and ADS modes respectively. Our simulation results and experimental data agree well, suggesting that our HuMADS framework is capable of replicating driver and vehicle performance during vehicle overtaking across our three defined modes.



\subsection{Simulation study for driver-ADS shared control }


To investigate the effects of various driver-ADS conditions on human driver workloads and resulting driving performance, we conduct a simulation study using the HuMADS framework incorporated with the ADS in an overtaking vehicle task. 

\paragraph{Evaluated Conditions}
The following driver-ADS conditions are considered based on the presence of ADS support and type of conflict:
\begin{enumerate}
    \item Manual control (\textbf{MC}): This is equivalent to the SAE autonomous Level 0 scenario where the vehicle is in full control of the driver without ADS support. This is considered as the baseline condition.
    \item Haptic shared control (\textbf{No Conflict}): This is equivalent to the SAE autonomous Level 2 scenario, where the driver model and the ADS share the steering task with no conflict, i.e. both agents have a consistent goal of performing the overtaking maneuver and same resulting trajectory.
    \item Haptic shared control with Type I Conflict (\textbf{Conflict-I}): In this condition, we simulate a Type I conflict scenario where both driver model and ADS share the steering task, but have a mismatch in their planned maneuver as described in Section IV.A. 
    \item Haptic shared control with Type II Conflict (\textbf{Conflict-II}): In this condition, we simulate a Type II conflict scenario where there is a mismatch in the resulting trajectory to achieve the overtaking task. Here, we modify the driving task reasoning parameters of the ADS by setting the steering compensation gain $h$ = 1.5 and time delay $\tau_{sw}$ = 0.05. This makes the controller more sensitive to track the desired lateral trajectory resulting in a more aggressive overtaking maneuver.
\end{enumerate}

In the study, we consider the two human driver model modes (\textbf{Tense} and \textbf{Relaxed}) for each condition.

\vspace{3mm}

\paragraph{Evaluation Metrics} The following measures were expressed to evaluate the effects of different HSC conditions on driver performance and control workload:
\begin{enumerate}
    \item Driving performance measures: Peak value of the lateral trajectory error, which is the difference between the input lateral trajectory and the actual lateral trajectory from the HuMADS simulation.
    \item Force-Torque measures: (i) Steering torques generated by the ADS, (ii) Resulting steering torque measured on the steering column between the driver model and the ADS actuator (see Fig.~\ref{fig:virtualmarkers}), (iii) Reaction point-to-point force on the driver's right hand.
    \item Control workload measures: (i) Joint actuation effort - estimated as the time integrated join torque, (ii) Control stress - estimated as the peak magnitude of the difference between the joint torque and the average nominal joint torque, (iii) Control load quantity - is estimated as the time integrated control stress.
\end{enumerate}

\begin{figure}[]
    \centering
    \includegraphics[width=0.975\columnwidth]{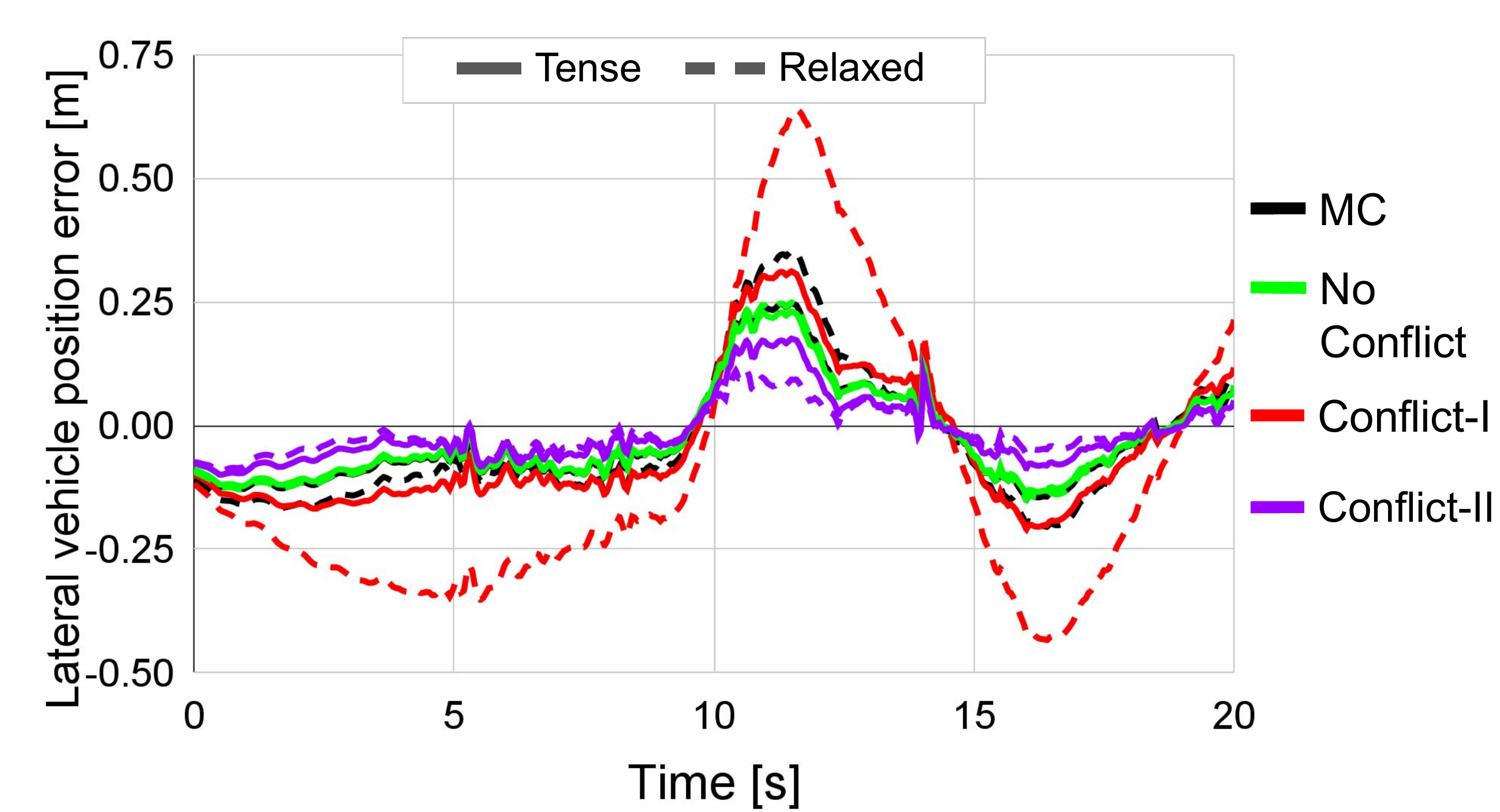}
    \caption{Comparisons of vehicle lateral position errors.}
    \label{fig:lateral_errors}
\end{figure}

\begin{figure}[]
    \centering
    \includegraphics[width=0.976\columnwidth]{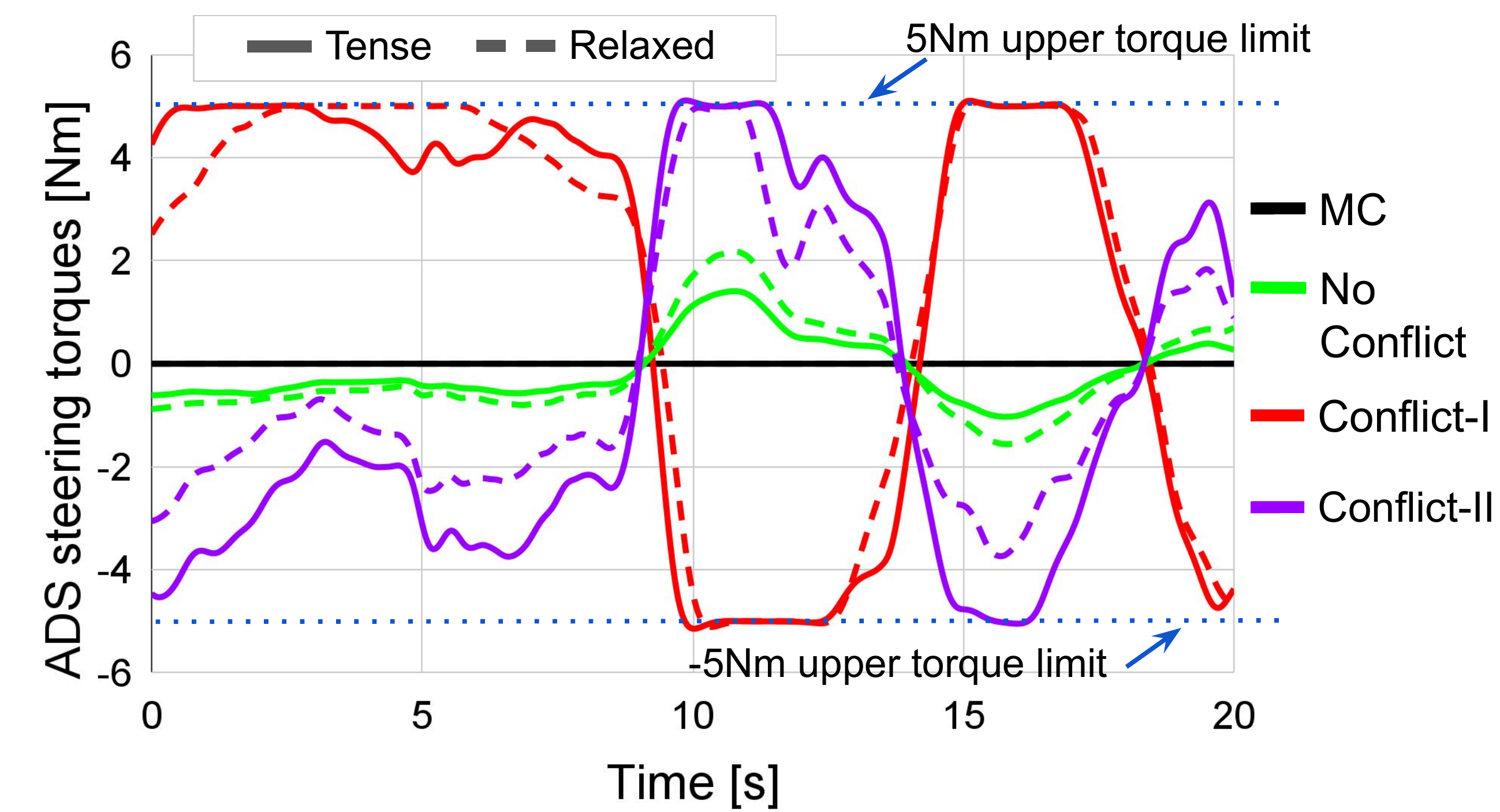}
    \caption{Steering actuation torques generated by the ADS.}
    \label{fig:actuated_torques}
\end{figure}

\begin{figure}[]
    \centering
    \includegraphics[width=0.975\columnwidth]{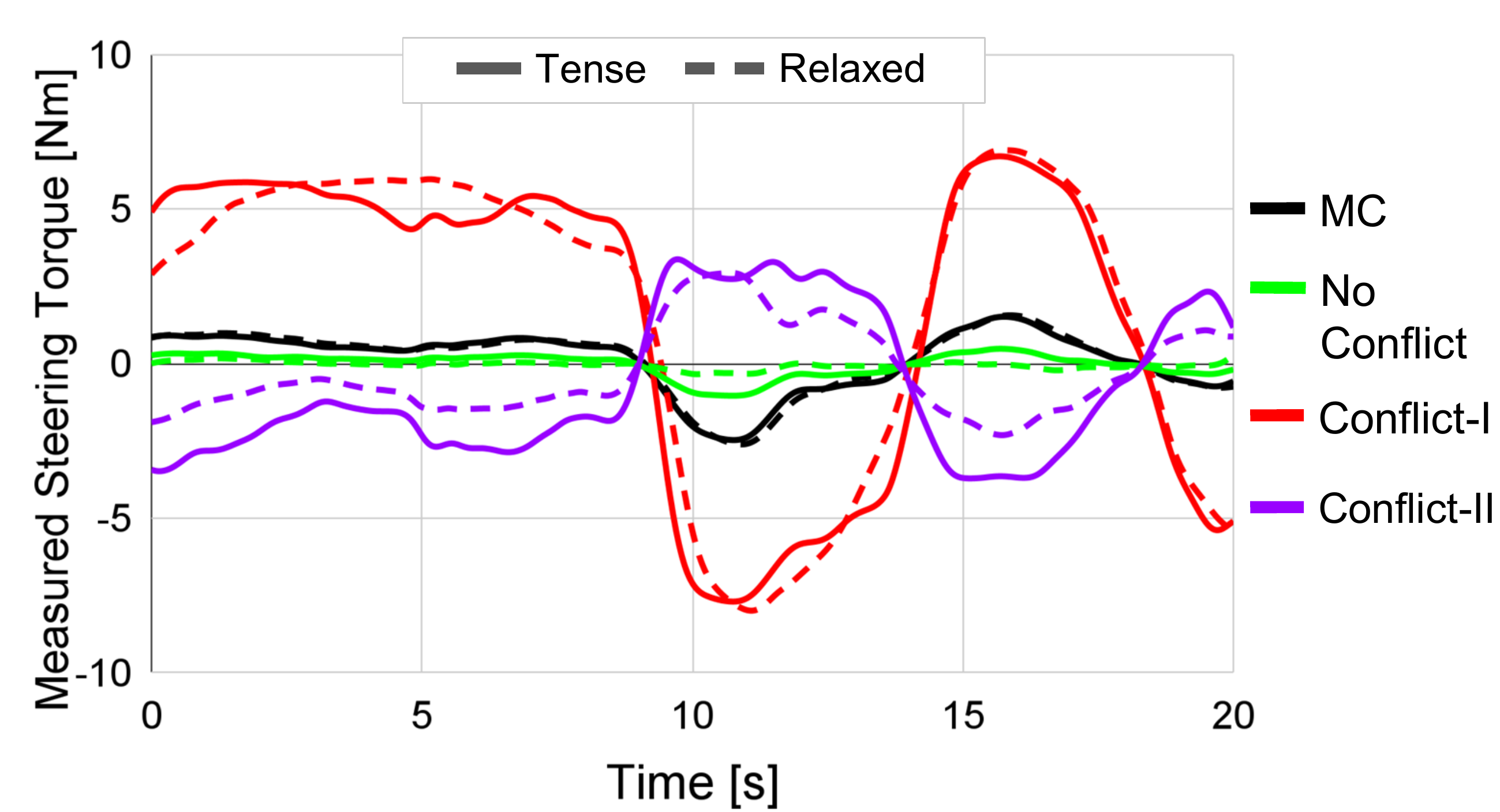}
    \caption{Steering torques measured by torque sensor on the steering column.}
    \label{fig:torque_measured}
\end{figure}

\begin{figure}[]
    \centering
    \includegraphics[width=0.975\columnwidth]{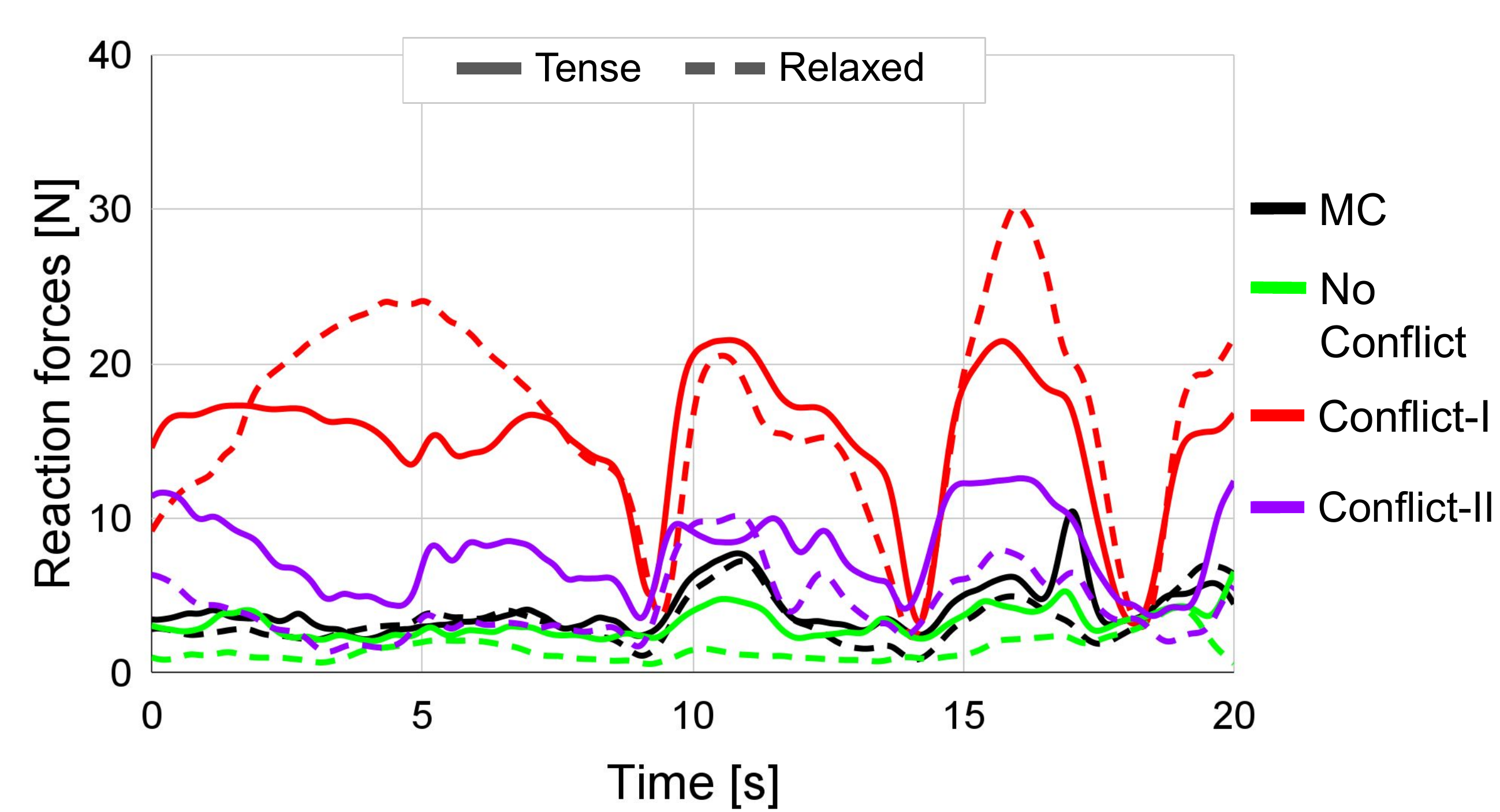}
    \caption{Reactions of point-to-point force on right hand.}
    \label{fig:reaction_forces}
\end{figure}

\begin{figure}[]
    \centering
    \includegraphics[width=0.975\columnwidth]{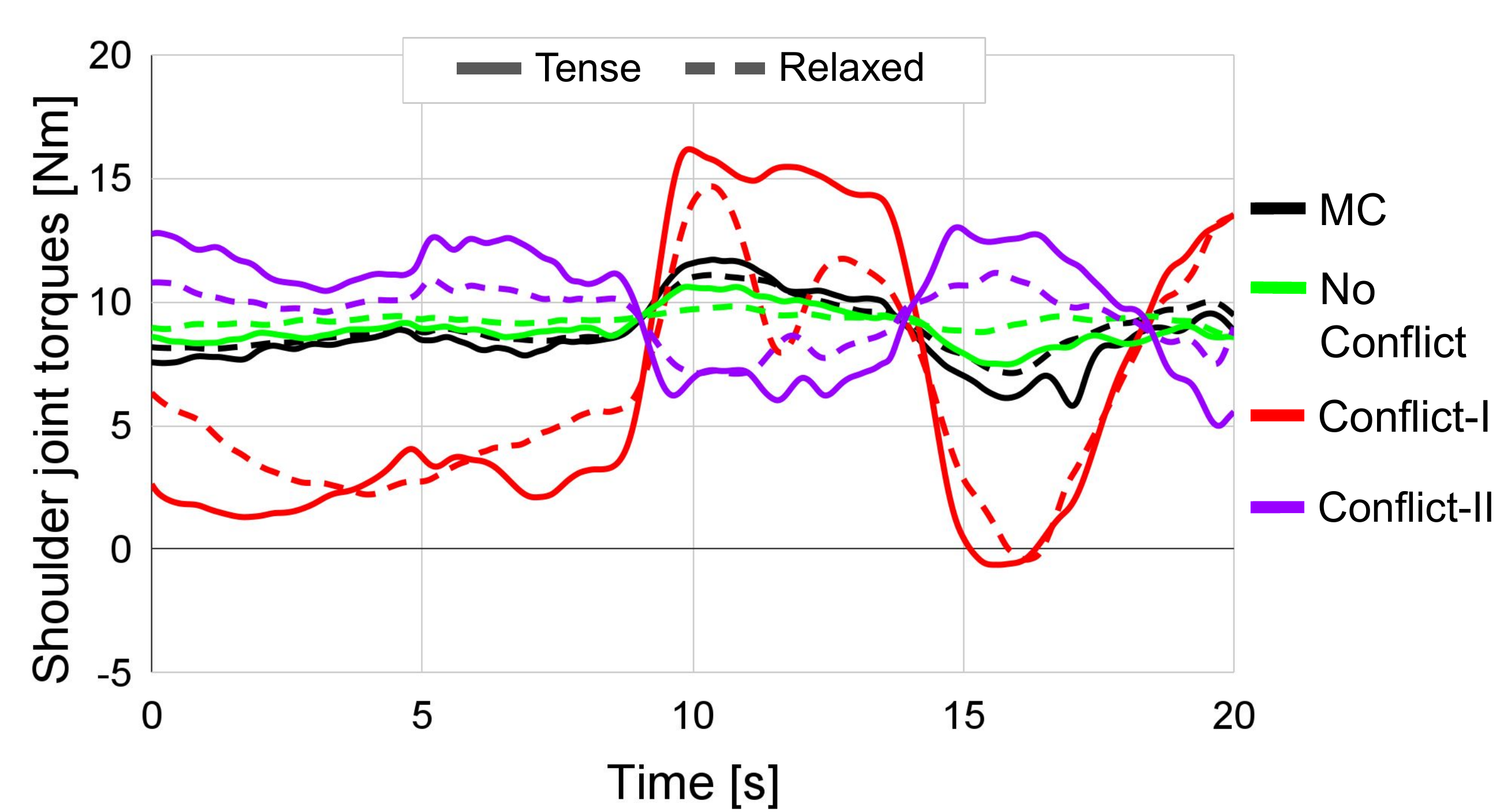}
    \caption{Actuated torques on the human driver shoulder joint on flexion and extension axis.}
    \label{fig:shoulder_torques}
\end{figure}

\begin{figure}[]
    \centering
    \includegraphics[width=0.975\columnwidth]{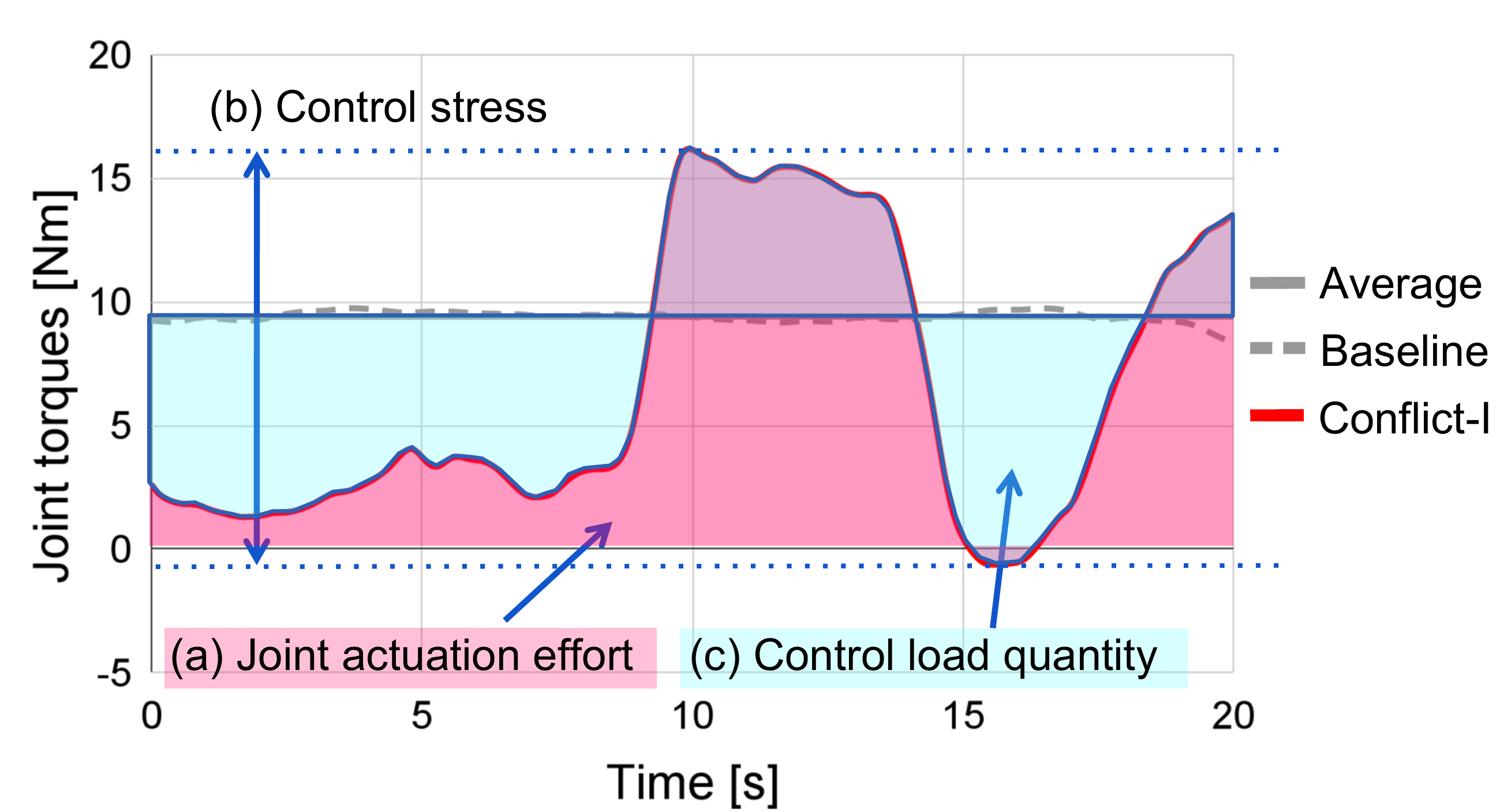}
    \caption{Estimated workload indicators for the shoulder joint of human driver during steering maneuver.}
    \label{fig:workload_measures}
\end{figure}

\section{Simulation Results}

\begin{table*}[]
\small
\centering
\caption{Estimated control workload values on shoulder and elbow joints during a 20 s simulated overtaking task. The numbers in parentheses indicate the descending order.}
\begin{tabular}{cccccccccc}
\multicolumn{1}{l}{} & \multicolumn{1}{l}{}        & \multicolumn{8}{c}{\textbf{Conditions}}                                                                                                                          \\ \cline{3-10} 
\multicolumn{1}{l}{} & \textbf{}                   & \multicolumn{2}{c}{\textbf{MC}}   & \multicolumn{2}{c}{\textbf{No Conflict}}  & \multicolumn{2}{c}{\textbf{Conflict-I}} & \multicolumn{2}{c}{\textbf{Conflict-II}} \\ \cline{3-10} 
\multicolumn{1}{l}{} & \textbf{Workload Indicator} & \textbf{Tense} & \textbf{Relaxed} & \textbf{Tense} & \textbf{Relaxed} & \textbf{Tense}      & \textbf{Relaxed}      & \textbf{Tense}      & \textbf{Relaxed}     \\ \hline \hline
Right shoulder       & Actuation effort            & 174            & 180              & 179            & 185              & 184                 & 142                   &  201                 & 190                  \\
                     & {[}Nm s{]}                  & (7)            & (5)              & (6)            & (3)              & (4)                 & (8)                   & (1)                 & (2)                  \\ \cline{2-10} 
                     & Control stress              & 5.9            & 4.0              & 3.1            & 1.1              & 16.8                & 15.1                  & 8.0                 & 4.1                  \\
                     & {[}Nm{]}                    & (4)            & (6)              & (7)            & (8)              & (1)                 & (2)                   & (3)                 & (5)                  \\ \cline{2-10} 
                     & Control load quantity       & 27.0           & 18.1             & 15.3           & 4.1              & 119.0               & 88.3                  & 45.4                & 19.6                 \\
                     & {[}Nm s{]}                  & (4)            & (6)              & (7)            & (8)              & (1)                 & (2)                   & (3)                 & (5)                  \\ \hline
Right elbow          & Actuation effort            & 82             & 47               & 40             & 49               & 1335                & 54                    & 49                  & 51                   \\
                     & {[}Nm s{]}                  & (2)            & (7)              & (8)            & (5)              & (1)                 & (3)                   & (6)                 & (4)                  \\ \cline{2-10} 
                     & Control stress              & 5.4            & 1.4              & 2.6            & 0.8              & 8.0                 & 5.3                   & 2.8                 & 1.3                  \\
                     & {[}Nm{]}                    & (2)            & (6)              & (5)            & (8)              & (1)                 & (3)                   & (4)                 & (7)                  \\ \cline{2-10} 
                     & Control load quantity       & 19.5           & 6.5              & 14.0           & 2.2              & 57.6                & 30.1                  & 12.2                & 6.1                  \\
                     & {[}Nm s{]}                  & (3)            & (6)              & (4)            & (8)              & (1)                 & (2)                   & (5)                 & (7)                  \\ \hline
\end{tabular}
\label{tab:workload_values}
\end{table*}

Fig.~\ref{fig:lateral_errors} shows comparison of vehicle lateral position errors in the overtaking task across our conditions. This error is computed based on the experimental overtaking data~\cite{naranjo2008lane} set as the desired lateral trajectory into our model. The largest maximum error (0.64 m) was observed in the Conflict-I condition with a relaxed driver mode. The second largest max error (0.34 m) occurred in the manual control condition with a relaxed driver mode. The Conflict-II condition had the least maximum error for both tense (0.19 m) and relaxed (0.11 m) driver modes, with the No Conflict condition in the middle with 0.25 m for both tense and relaxed driver modes.

Fig.~\ref{fig:actuated_torques} shows the steering torques generated by the ADS $T_{ADS}$. The peak ADS torques were less than the maximum torque limit of 5 Nm only for the No Conflict condition (both Tense and Relaxed modes), while both conflict conditions (Conflict-I \& -II) became saturated at the torque limit. It is interesting that even the Conflict-II condition reached the torque limited where both the driver model and the ADS are driving in the same direction, however, with different resulting trajectories. 

Fig.~\ref{fig:torque_measured} shows the resulting steering torques measured on the steering column between the driver model and the ADS actuator. This value reflects the torque exerted by the driver model on the steering wheel $T_{hm}$. As anticipated, the largest peak reaction torque was observed with the Conflict-I condition followed by the Conflict-II condition suggesting that the human driver tries to exert more torque on the steering wheel to override the ADS torques. We observed that the driver steering torque was reduced from 2.47 Nm in the manual control condition to 1.04 Nm in the No Conflict condition. We observed the least driver steering torque in the No Conflict condition because both agents are in alignment, hence, the human performs less work to achieve the overtaking maneuver.

Fig.~\ref{fig:reaction_forces} shows reactions of point-to-point force on the right hand. The No Conflict with relaxed driver mode had the lowest peak reaction force (4.2 N) whereas the Conflict-I with relaxed driver mode had the highest peak reaction force (30.1 N). It is interesting to see that the No Conflict conditions resulted in less peak reaction force than even the manual control (MC) condition.

Fig.~\ref{fig:shoulder_torques} shows actuated torques on the human driver’s right shoulder joint on flexion and extension axis. From the plot, we observed that the human driver shoulder generated a certain nominal torque across all the conditions. The baseline trace is the shoulder joint torque (9.4 Nm) generated to keep the arm lifted against the force of gravity due to the gravity compensation component of our human motion control equation (3).

Fig.~\ref{fig:workload_measures} shows the estimated workload indicators for the shoulder joint for a representative set of case scenarios. TABLE~\ref{tab:workload_values} shows a summary of driver control workload during 20 s of overtaking task. The actuation effort, control stress, and control load quantity are indicated for the right shoulder and right elbow joints with the descending order of ranking numbers in the parenthesis. For the shoulder joint, the Conflict-II condition (both Tense and Relaxed) resulted in the highest actuation effort with the lowest coming from the MC with tense driving and Conflict-I with relaxed driving. We found that in some of the conditions, the relaxed driver mode resulted in higher actuation effort in the shoulder and elbow joints than the tense driver mode. This case was different for the control stress and load quantity. Under all conditions, relaxed human driver models showed less control stress and load quantity than tense mode. The Conflict-I condition showed significantly higher values of control stress and load quantity compared to the other conditions. Less control stress and load quantity were generally observed with the No Conflict condition. However, we observed that some values of Conflict-II were greater than those of the manual control condition.

\section{Discussion}

\subsection{Effects on driving performance}

Existing experimental user studies have shown that Level 2 haptic-based steering assistance improves driving performance~\cite{petermeijer2015effect}. Based on our simulation study, the No Conflict condition (both tense and relaxed) had a peak lateral position error of 0.25 m, equivalent to the manual control condition with tense driver, but less than the manual control condition with relaxed driver (0.35 m). The result suggests that a relaxed driver using a well-tuned ADS (i.e. No Conflict condition) can achieve the same steering accuracy as a tense driver in manual control, thereby reducing the steering effort required.

Figure 9 shows the lowest peak lateral errors in the Conflict-II case (both relaxed and tense modes). Tracking performance improvements can be attributed to the ADS's task reasoning model being set at a more sensitive level. Also, the tense driving by the driver model increased the lateral error compared to the relaxed driving mode. This is consistent with literature~\cite{mulder2008effect} showing that haptic assistance improved driving performance, but with the trade-off of increased steering forces due to misaligned resulting trajectories. Research on assistance personalization attempts to develop more aligned task reasoning models of the human driver to resolve type II conflicts and improve driver acceptance and satisfaction while using ADAS~\cite{boink2014understanding}. We propose that our simulation framework can be used to test and validate the haptic shared control systems while still in the system design phase.

As expected, the Conflict-I condition, specifically when the driver is tense, results in the largest peak lateral error (0.64 m). This error, which is caused by a struggle between the driver and the ADS, can lead to bad driving performance and unpredictable maneuvers with other road vehicles, resulting in accidents. Muslim and Itoh~\cite{muslim2018effects} studied an example of Conflict-I where the human driver intends to make an overtaking maneuver but fails to recognize that an oncoming vehicle is in their blind spot, which the ADS can see. This scenario is fairly common, and other techniques for alerting the driver to oncoming vehicles in their blind spots, including visual or audio warning signals, haptic vibrations, etc. have been proposed~\cite{hoc2009cooperation}.

\subsection{Expected actuated joint torques on the human driver}

The peak actuated joint torques of the shoulder and elbow joints were 19.3 Nm and 9.3 Nm, respectively. An experiment measuring joint torques of human participants in a driving scenario showed that maximum flexion torques at the shoulder and elbow joints range between 80 Nm and 50 Nm, respectively~\cite{choi2005experimental}. Therefore, the predicted joint torques by the HuMADS are within the range of human actuation. However, although the actuated torques are within the range of human actuation, further verification of equilibrium joint torques in steering control scenarios are necessary. Our study utilized robot manipulation control equations, including gravity compensation, to compute joint torques, but it is possible that drivers rest some or all of their arm weight on the steering wheel rather than holding them up with their own joint actuation. Additionally, seat posture and interactions with the armrest, seat, and other body parts of the human driver may influence the effort required to hold the steering wheel.   


\subsection{Effect on human driver joint loading and estimated physical workload}

Through the HuMADS simulation framework, we were able to compare the effects of various shared control conditions on both joint loading and estimated control workload for human drivers. The majority of previous studies have used human driver subjective evaluations~\cite{hosseini2016predictive, de2014effects} or highly instrumented experimentation~\cite{nacpil2021application}, making it cumbersome to estimate actual physical human workloads from measurable parameters. 
From Table~\ref{tab:workload_values}, we observe that Conflict-I condition led to the highest values across most of the physical workload indicators (except the shoulder actuation effort). This is consistent with experimental studies. Mulder et al.~\cite{mulder2008effect} found that haptic support may lead to increased driver steering torques due mismatches in the shared goal which often results in higher control workloads. Fig. 10-12 show greater values for reaction force and torque values for Conflict-I condition compared to other conditions.

Compared to Conflict-I, the Conflict-II condition led to a lower workload value. While Conflict-II condition resulted in improved driving performance, the trade-off was increased control workload over an ADS with well-tuned task reasoning model (i.e. No Conflict condition). A similar study by De Jonge et al.~\cite{de2015effect} showed that adapting the haptic shared controller to individual operators reduced trajectory conflicts and shared control forces. In our analysis, we also found that, across most of the control workload measures, the No Conflict condition with relaxed driver mode led to the lowest values as well as the best driving performance. By using our proposed framework, it is possible to analyze the effect of data-driven adaptive shared control systems on control workloads and driving performance in a variety of driving situations without needing more elaborate, cumbersome experimental studies.



\section{Limitations and Future Work}

This study employed tense and relaxed human driver modes from validation results with Pick and Cole’s study~\cite{pick2007dynamic}. The control gains were set as constant parameters to express two different human driver modes. However, real human drivers may have more adaptive neuromuscular response models in accordance with visual, acoustic, and haptic feedback. Additionally, other factors affect driver behavior and driving performance such as driver cognitive state and alertness (e.g., drowsiness, fatigue, distraction, etc.) which has not been considered in the HuMADS framework. To develop comprehensive surrogate models of real human test drivers in simulation, further modeling based on human science is needed. 

Also, this study considered conflicts arising from mismatches in the maneuver decision level and in the resulting trajectory, however, Abbink et al.~\cite{abbink2012importance} has shown the importance of ADS to accurately model and respond to the neuromuscular dynamics of drivers in mitigating conflicts and disagreements in haptic shared driving scenarios. Future work will explore this direction on our HuMADS framework. Furthermore, although our findings generally agree with existing experimental studies, there is need to further validate the control workload measures experimentally with a human user study.

\section{Conclusion}


The paper proposes a novel method of evaluating the effects of Level 2 driver assistance systems on the estimated driver workload, force-torque interactions and driving performance. Our approach uses the human model-based active driving system (HuMADS) proposed in~\cite{kimpara2019human} to simulate realistic driver-vehicle interaction during driving tasks. A vehicle overtaking task was used in this study to evaluate the effects of haptic shared control on the simulated driver model. Equivalent tense and relaxed arm-steering impedance models were developed and validated against human participant data. Additionally, the vehicle overtaking performance generated by the HuMADS was validated using experimental road driving data.

In line with previous human studies, we found a significant impact of control conflicts on driving performance, force-torque loads, and control workload in the driver model. Despite both tense and relaxed human driver models completing overtaking tasks successfully, the No conflict shared control mode improved overtaking performance, especially for relaxed drivers. Additionally, Type I conflict conditions had the highest workload followed by Type II conflict conditions. Although our simulation results generally agree with those of existing work, improvements need to be made to the model to increase bio-fidelity, and further experimental validation of internal dynamic variables is needed. We believe that this framework will aid future research on driver-ADS interaction in Level 2 systems, improving their design, safety, and overall usability~\cite{toyota-teammate}.



\section{Acknowledgment}

We would like to thank team members of the National Center for Simulation in Rehabilitation Research (NCSRR) for providing extensive technical support on the OpenSim API and simulator.



\ifCLASSOPTIONcaptionsoff
  \newpage
\fi



\bibliographystyle{IEEEtran}

\bibliography{references}
\begin{IEEEbiography}
[{\includegraphics[width=1in,height=1.25in,clip,keepaspectratio]{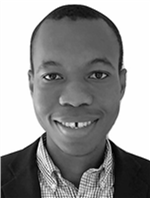}}]
{Kenechukwu C. Mbanisi}
is an assistant professor of robotics engineering at Olin College of Engineering. He received his B.Eng. degree in electrical and electronic engineering from Covenant University, Nigeria, in 2013, an M.S. degree in 2018, and a Ph.D. in 2022, both in robotics engineering from Worcester Polytechnic Institute, Massachusetts, USA. His research experience interests include human-robot interaction, shared autonomy in (semi-)autonomous systems, haptics, social robot navigation, and STEM/robotics education.

\end{IEEEbiography}

\begin{IEEEbiography}
[{\includegraphics[width=1in,height=1.25in,clip,keepaspectratio]{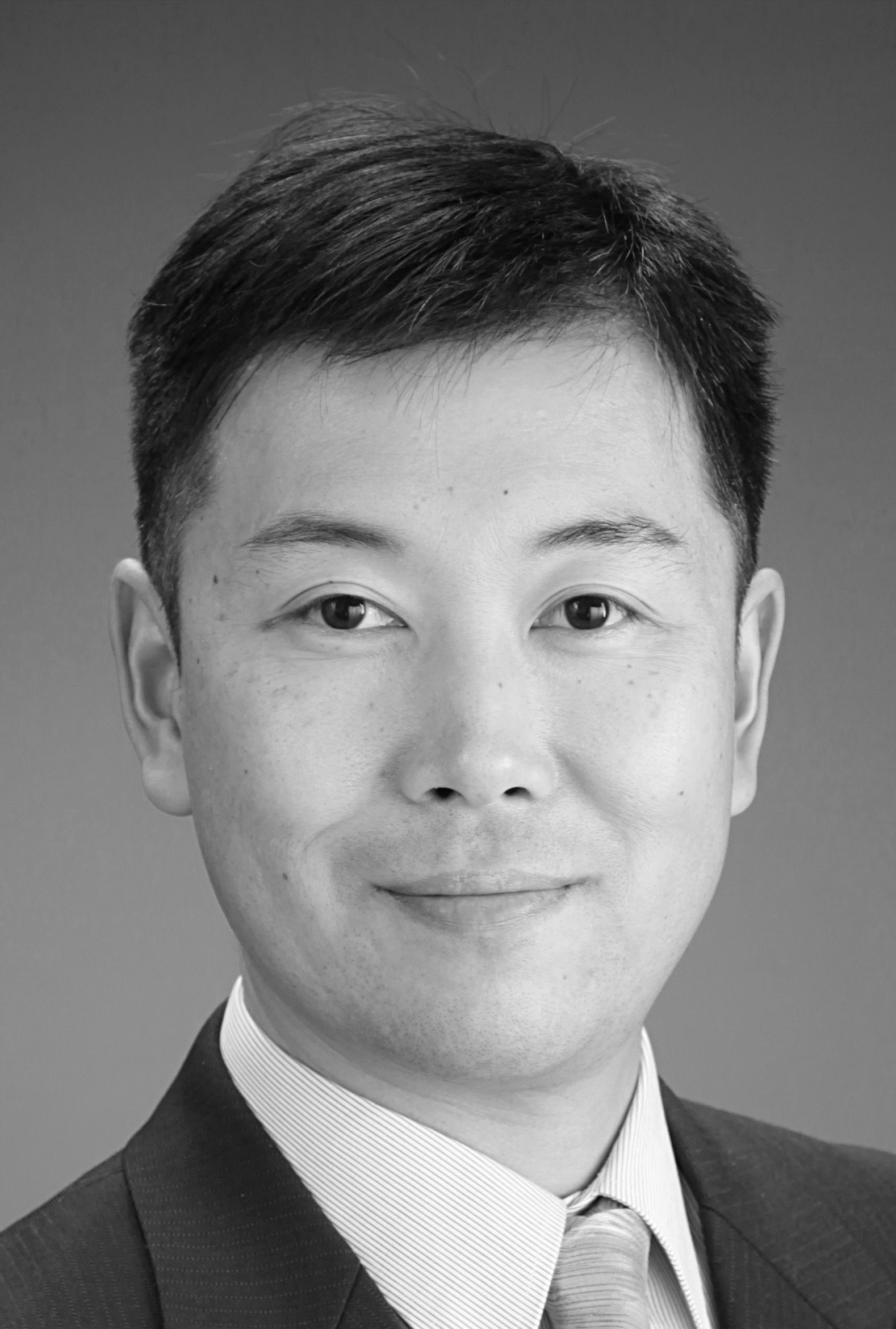}}]
{Hideyuki Kimpara}
received the B.S. and M.S. degrees from Keio University, Japan, in 1998 and 2000, respectively, and the Dr.Eng. degree from Nagoya University, Japan, in 2015. In 2000, he joined Toyota Central R\&D Labs., Inc., Japan. While serving with injury biomechanics research for small females and rib fracture injury studies, he visited the Bioengineering Center of Wayne State University, Detroit, MI, from Aug 2002 to Jan 2006. In 2016, he was dispatched to Toyota Research Institute North America (TRINA), Ann Arbor, MI, and assigned human motion research with Worcester Polytechnic Institute, Worcester, MA. Since 2020, he has been involved in human augmentation research for a safer and more prosperous life at Kyocera Minato-Mirai Research Center, Japan. His current research interests include human motor skills, robotics, and applied biomechanics.
\end{IEEEbiography}

\begin{IEEEbiography}
[{\includegraphics[width=1in,height=1.25in,clip,keepaspectratio]{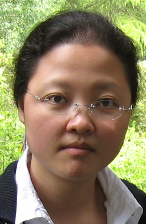}}]
{Zhi Li}
is an assistant professor with Robotics Engineering Program and Mechanical Engineering Department at Worcester Polytechnic Institute. She received B.S. degree in mechanical engineering from China Agricultural University, Beijing, China, in 2006, M.S. degree in mechanical engineering from University of Victoria, Victoria, BC, Canada in 2009, and Ph.D. in computer engineering from University of California, Santa Cruz in 2014. She was a Postdoctoral Associate with electrical and computer engineering at Duke University from 2015-2016. Her research experience covers upper limb stroke rehabilitation, multi-arm surgical robot system, and humanoid nursing robot. Since she joined in WPI in 2017, she primarily focuses on modeling and learning of human motion coordination, and take human-shared-autonomous robot motion coordination. 
\end{IEEEbiography}

\begin{IEEEbiography}
[{\includegraphics[width=1in,height=1.25in,clip,keepaspectratio]{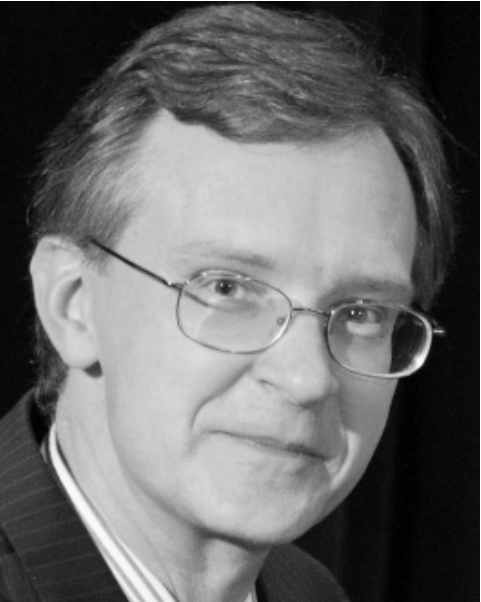}}]
{Danil Prokhorov}
(SM’02) received the M.S. degree
(Hons.) in Saint Petersburg, Russia, in 1992, and the Ph.D. degree in 1997. He joined the Staff of the Ford Scientific Research Laboratory, MI, USA. While at
Ford, he pursued machine learning research focusing on neural networks with applications to system modeling, powertrain control, diagnostics, and optimization. He has been involved in research and planning for various intelligent technologies, such as highly automated vehicles, AI and other futuristic
systems at the Toyota Tech Center, Ann Arbor, MI, USA, since 2005. Since 2011, he has been the Head of the Future Research
Department, Toyota Motor North America R\&D. He started his research career
in Russia. He studied system engineering which included courses in math,
physics, mechatronics, computer technologies, and aerospace and robotics.
He has been serving as a Panel Expert for NSF, DOE, ARPA, as a Senior
and Associate Editor for several scientific journals for more than 20 years.
He has been involved with several professional societies, including the IEEE
Intelligent Transportation Systems and the IEEE Computational Intelligence,
and the International Neural Network Society (INNS) as its former Board
Member, the INNS President, and was recently elected as an INNS Fellow.
He has authored lots of publications and patents. Having shown feasibility
of autonomous driving and personal flying mobility, his department continues
research of complex multi-disciplinary problems while exploring opportunities
for the next big thing. 
\end{IEEEbiography}

\begin{IEEEbiography}
[{\includegraphics[width=1in,height=1.25in,clip,keepaspectratio]{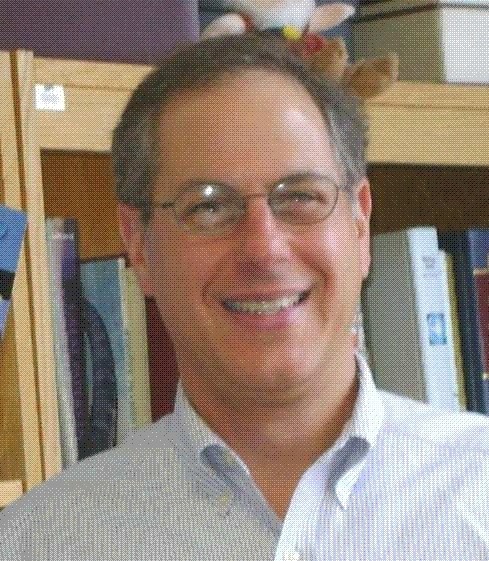}}]
{Michael A. Gennert}
(SM‘11) is Professor Emeritus of Robotics Engineering, CS, and ECE at Worcester Polytechnic Institute, where he directed the WPI Humanoid Robotics Laboratory and was Founding Director of the Robotics Engineering Program. He has worked at the University of Massachusetts Medical Center, the University of California Riverside, PAR Technology Corporation, and General Electric. He received the S.B. in CS, S.B. in EE, and S.M. in EECS in 1980 and the Sc.D. in EECS in 1987 from MIT.  Dr. Gennert's research interests include robotics, computer vision, and image processing, with ongoing projects in humanoid robotics, robot navigation and guidance, biomedical image processing, and stereo and motion vision.  He led WPI teams in the DARPA Robotics Challenge and NASA Space Robotics Challenge. He is author or co-author of over 100 papers.  His research has been supported by DARPA, NASA, NIH, NSF, and industry. He is a member of Sigma Xi, and a senior member of IEEE and ACM.
\end{IEEEbiography}



\end{document}